\documentstyle[a4]{article}

\title{\bf Tagging the Teleman Corpus}

\author{{\bf Thorsten Brants and Christer Samuelsson}\\
	Universit\"at des Saarlandes\\
	FR 8.7, Computerlinguistik, Postfach 151150\\
	D-66041 Saarbr\"ucken, Germany\\
	Internet: {\tt \{thorsten,christer\}@coli.uni-sb.de}}

\date{\em In Proceedings of the 10th Nordic Conference of\\
	Computational Linguistics, Helsinki, Finland, 1995.}

\begin{document}

\maketitle

% ------------------------------------------------------------------- Abstract

\section*{\centering Abstract}

\begin{quote}
Experiments were carried out comparing the Swedish Teleman and the
English Susanne corpora using an HMM-based and a novel reductionistic
statistical part-of-speech tagger. They indicate that tagging the
Teleman corpus is the more difficult task, and that the performance of
the two different taggers is comparable.
\end{quote}

% ---------------------------------------------------------------------- Intro

\section{Introduction}

The experiments reported in the current article continue a line of
research  in the field of part-of-speech tagging using self-organizing
models that was presented at the previous (9th) Scandinavian Conference
on Computational Linguistics. Then, the well-established HMM-based Xerox
tagger, see \cite{Cutting:94}, was  compared with some less known
taggers, namely a neural-network tagger described in
\cite{Eineborg;Gamback:94}, and a Bayesian tagger presented  in
\cite{Samuelsson:94}. The Xerox tagger performs lexical generalizations
by clustering words based  on their distributional patterns, while the
latter two utilize the morphological information present in Swedish by
generalizing over word suffixes.

This time, another HMM-based approach is compared with a novel
reductionistic statistical tagger inspired by the successful Constraint
Grammar  system, \cite{KarlssonEA:95}.

The performed experiments do not only serve to evaluate the two taggers,
but also shed some new light on the Teleman corpus as an evaluation
domain for part-of-speech taggers compared to other, English, corpora.

The paper is organized as follows: Section \ref{Teleman} discusses the
Teleman corpus and the tagsets used. Section~\ref{HMM} describes the
HMM-based tagger  and Section~\ref{RedStat} the reductionistic
statistical one. The vital issue of handling sparse data is addressed in
Section~\ref{Sparse} and the experimental results are presented in
Section~\ref{Exp}.

% ------------------------------------------------------------- Teleman Corpus

\section{The Teleman Corpus}
\label{Teleman}

The Teleman corpus \cite{Teleman:74} is a corpus of contemporary
Swedish, representing a mixture of different text genres like
information brochures on military service and medical care, novels, etc.
It comprises 85,408 words (tokens; here, words is a collective
denotation of proper words, numbers, and punctuation). There are 14,191
different words (types); the most frequent one is ``.'', which occurs
4,662 times; the most frequent proper word is ``{\em och\/}'' ({\em
and\/}), which occurs 2,217 times. 8,458 of the words occur exactly
once, which is 60\% of the types but only 10\% of the tokens.

For the experiments, we used two different tagsets. First, we used the
original tagset, consisting of 258 tags. Each of the 14,191 word types
can have between one and 15 of the 258 tags (the highly ambiguous word
``{\em f\"or\/}'' ({\em for, stern, lead, too, \dots}) has the maximum
number of tags). We then used a reduced tagset, consisting of 19 tags,
which represent  common syntactic categories and punctuation. This
tagset is identical  to that used in the publications mentioned above.
Each of the word types then has between one and 7 tags (``{\em
f\"or\/}'' and ``{\em i\/}'' have the maximum number of tags).

% -------------------------------------------- Teleman Corpus ----- Comparison

\subsection{Comparison with an English Corpus}

Since 10\% of the words in the Teleman corpus occur only once, we expect
from the Good-Turing formula \cite {Good53} that 10\% of the words in
new text be unknown, which is a very high percentage. Other publications
typically report 5\%. Since most of the work in this area is on English
corpora, we compared the Teleman corpus with an English corpus, namely
the Susanne corpus \cite{Sampson95}, which is a re-annotated part of the
Brown Corpus \cite{Francis82}, comprising different text genres. The
relevant facts are summarized and compared in Table~\ref{CompareTable}.
The major difference (apart from corpus size and tagsets used) is the
percentage of words that occur exactly once: 10\% for Teleman vs.\ 4\%
for Susanne. According to the Good-Turing formula, this percentage is
identical to the expected percentage of unknown words. Actual counts by
dividing the corpora into training and test parts yield around 14\% and
7\%, respective. This indicates that unseen Swedish text will have
substantially more unknown words than unseen English, which is most
likely due to the higher degree of morphological variation in Swedish.

A further difficulty with the Swedish corpus is the higher degree of
ambiguity. In the Teleman corpus, each word in the running text has  in
average 2.38 tags for the small tagset, and 3.69 for the large tagset.
These numbers are 2.07 and 2.61 for the Susanne corpus, despite the fact
 that the tagsets for the Susanne corpus are larger than those for the
Teleman corpus.  Thus, there is much more work for the tagger to do in
the Teleman corpus. Some more numbers: in the running text,
54.5\%/64.2\% of the words in the Teleman corpus are ambiguous, and only
44.3\%/48.9\% in the Susanne corpus (small/large tagset, resp.; see
Table~\ref{AmbiguityTable} for further details).

\begin{table}
\hrule
\caption{Comparison of Teleman and Susanne corpora}
\label{CompareTable}
\begin{center}\small\vspace*{-3ex}
\begin{tabular}{@{\hspace{1pt}}l|rr@{\hspace{1pt}}}
		& Teleman	& Susanne \\
\hline
size		&  85,408 words	& 156,644 words\\
word types	&  14,191 words	&  14,732 words\\
most freq.~word	& 4662$\times$``.'' & 9641$\times$``{\em the\/}'' \\
one occurrence	& 8,458 words	& 6,820 words \\
unknown words	& 10\% expected & 4\% expected \\
\hline
\stepcounter{footnote}
tagset		&  258	tags	& 424 tags$^{\arabic{footnote}}$ \\
max.~tags/word	& 15 (for ``{\em f\"or\/}'') & 14 (for ``{\em as\/}'') \\
\hline
reduced tagset	&   19 tags	&  62 tags\\
max.~tags/word	& 7 (``{\em f\"or\/}'', ``{\em i\/}'') & 6 (``{\em
a\/}'', ``{\em no\/}'') \\
\end{tabular}
\end{center}
\hrule
$^{\arabic{footnote}}${\footnotesize Tags in the Susanne corpus with
indices are counted as separate tags.}
\end{table}

\begin{table}
\hrule
\caption{Distribution of number of categories per word in running text
for the Teleman and Susanne corpus, small and large tagsets.}
\label{AmbiguityTable}
\begin{center}\small
\begin{tabular}{l|rr|rr}
	& \multicolumn{2}{c|}{Teleman} & \multicolumn{2}{c}{Susanne} \\
	& small	& large & small & large \\
\hline
1	& 45.5\%& 35.8\%& 55.7\%& 51.1\%\\
\hline
2	& 16.2\%& 12.2\%& 17.4\%& 19.2\%\\
3	& 17.7\%& 14.1\%&  4.8\%&  4.2\%\\
4	&  6.4\%& 10.0\%& 11.1\%&  4.5\%\\
5	&  9.0\%&  5.3\%&  8.4\%&  9.2\%\\
6	&  1.7\%&  6.2\%&  2.6\%&  2.2\%\\
7	&  3.5\%&  4.3\%& --	&  2.2\%\\
8	& --	&  1.1\%& --	&  4.8\%\\
9	& --	&  2.9\%& --	& --	\\
10	& --	&  2.7\%& --	&  2.0\%\\
11	& --	&  1.6\%& --	& --	\\
12	& --	&  2.9\%& --	& --	\\
13	& --	& --	& --	& --	\\
14	& --	& --	& --	&  0.6\%\\
15	& --	&  0.9\%& --	& --	\\
\hline
$> 1$	& 54.5\%& 64.2\%& 44.3\%& 48.9\%\\
\end{tabular}
\end{center}
\hrule
\end{table}

\begin{table}
\hrule
\caption{Teleman corpus parts}
\label{TelemanPartsTable}
\begin{center}
\begin{tabular}{l|rr}
\stepcounter{footnote}
	& total words	& unknown words$^{\arabic{footnote}}$	\\
\hline
part A	& 67,402	& --- \\
part B	&  9,262	& 1,421 (15.3\%) \\
part C	&  8,774	& 1,198 (13.7\%) \\
\hline
$\Sigma$& 85,408	& \\
\end{tabular}
\end{center}
\hrule
$^{\arabic{footnote}}${\footnotesize Unknown words are words that
occur only in the test set, but not in the training set.}
\end{table}

\begin{table}
\hrule
\caption{Susanne corpus parts}
\label{SusannePartsTable}
\begin{center}
\begin{tabular}{l|rr}
	& total words	& unknown words	\\
\hline
part A	& 127,385	& --- \\
part B	&   9,752	& 714 (7.4\%) \\
part C	&   9,684	& 563 (5.8\%) \\
\hline
\stepcounter{footnote}
$\Sigma$& 146,821$^{\arabic{footnote}}$ & \\
\end{tabular}
\end{center}
\hrule
$^{\arabic{footnote}}${\footnotesize The remaining 9,823 words of the
Susanne corpus were not used in the experiments.}
\end{table}

% --------------------------------------------------------------- HMM Approach

\section{The HMM Approach}
\label{HMM}

A Hidden Markov Model (HMM) consists of a set of states, a set of output
symbols and a set of transitions. For each state and each symbol, the
probability that this symbol is emitted by that state is given. Also, a
probability is associated with each transition between states (see
\cite{Rabiner89} for a good introduction). The transition probability,
and thus the probability of the following state, depends only on the
previous state for first order HMMs, or on $k$ previous states for HMMs
of $k$th order. HMM approaches to part-of-speech tagging make the
well-known assumption that the current category or part-of-speech of a
word depends only on the previous $(n-1)$ categories (Markov
assumption), thus they assume that natural language is a Markov process
of order $(n-1)$, which of course is not true, but a successful
approximation. $n = 3$ is chosen in most of the cases, resulting in a
trigram model (i.e., always working with a window of size 3), since it
yields the best compromise between size of corpora needed for training
and tagging accuracy. Furthermore, the current word (symbol) depends
only on the current category (state). Thus, instead of calculating and
maximizing $P(T_1\dots T_k \mid W_1\dots W_k)$, with $T_i$ tags and
$W_i$ words, which is impossible in all practical cases, one calculates
and maximizes
\begin{equation}
\label{MarkovEqn}
	\prod\limits_{i=1}^{k}P(T_i \mid T_{i-n+1}\dots T_{i-1})P(W_i \mid T_i)
\end{equation}
to find the best sequence of tags for a given sequence of words.

The parameters of an HMM can be estimated directly from a pretagged
corpus via maximum-likelihood estimation (MLE). But MLE sets a lot of
the transition probabilities to zero, and if one of the multiplied
probabilities in (\ref{MarkovEqn}) is zero, the product becomes zero,
leaving no means to distinguish between different products that contain
a zero probability. This results in poor estimates for the probabilities
of new sequences of words. This problem is addressed in Section
\ref{Sparse}.

Another way of estimating the parameters of an HMM is to use an untagged
corpus, a lexicon with parts-of-speech lists for the words and the
Baum-Welch algorithm \cite{Baum72}. This approach has the advantage of
avoiding the tedious work of manually annotating a corpus, but it
requires a sophisticated choice of initial biases, and generally, the
performance is worse than that achieved with annotated corpora.

When using an HMM for tagging, the system gets a string of words and has
to find the most probable sequence of tags that could have produced
the string of words. This is done with a dynamic programming method, the
Viterbi algorithm \cite{Viterbi67}.
The algorithm finds the most probable sequence of states in time linear
in the length of the input string.

% ---------------------------------------------------- Reductionistic Approach

\section{The Reductionistic Statistical Approach}
\label{RedStat}

Although not yet fully realized, the basic philosophy behind the
reductionistic statistical approach is to give it the same expressive
power as the Constraint Grammar system.

\subsection{Constraint Grammar}
\label{CG}

The Constraint Grammar system performs remarkably well;
%\cite{Voutilainen+Heikkila:94}% wrong automatic linebreak
[Voutilainen \& Heikkil\"a 1994] report 99.7\% recall, or 0.3\%
error rate,  which is ten times smaller than that of the best
statistical taggers. These impressive results are achieved by:
\begin{enumerate} \item Utilizing a number of different information
sources, and not only the stereotyped lexical statistics and $n$-gram tag
statistics that have become the de facto standard in statistical
part-of-speech tagging. \item	 \label{YXA} Not fully resolving all
ambiguities when this would jeopardize the recall. \end{enumerate}
Property \ref{YXA} means that the system trades precision for recall,
which makes it ideal as a preprocessor for natural language systems
performing deeper analysis.

The Constraint Grammar system works as follows:
First, the input string is assigned all possible tags from the lexicon, or
rather, from the morphological analyzer.
Then, tags are removed iteratively by repeatedly applying a set of rules,
or constraints, to the tagged string.
When no more tags are removed by the last iteration, the process terminates,
and morphological disambiguation is concluded.
Then a set of syntactic tags are assigned to the tagged input
string and a similar process is performed for syntactic disambiguation.
This method is often referred to as {\it reductionistic tagging\/}.

The rules are sort-of formulated as finite state automata [Tapanainen,
personal communication], which allows very fast processing.

Each rule applies to a current word with a set of candidate tags.
The structure of a rule is typically:
\begin{quote}
``In the following context, discard the following tags.''
\end{quote}
or
\begin{quote}
``In the following context, commit to the following tag.''
\end{quote}
We will call discarding or committing to tags the {\it rule action\/}.
A typical {\it rule context\/} is:
\begin{quote}
``There is a word to the left that is unambiguously tagged with the
following tag, and there are no intervening words tagged with such
and such tags.''
\end{quote}

\subsection{The New Approach}
\label{NewRed}

The structure of the Constraint Grammar rules readily allows their contexts
to be viewed as the conditionings of conditional probabilities,
and the actions have an obvious interpretation as the corresponding
probabilities.

Each context type can be seen as a separate information source, and we will
combine information sources $S_1,\ldots,S_n$ by multiplying the scaled
probabilities:
\begin{eqnarray*}
\frac{P(T \mid S_1,\ldots,S_n)}{P(T)} \;\; \approx \;\;
\prod_{i=1}^{n} \frac{P(T \mid S_i)}{P(T)}
\end{eqnarray*}
This formula can be established by Bayesian inversion, then performing the
independence assumptions, and renewed  Bayesian inversion:
\begin{eqnarray*}
\lefteqn{P(T \mid S_1,\ldots,S_n) \;\; =}\\
&=& \frac{P(T) \cdot P(S_1,\ldots,S_n \mid T)}{P(S_1,\ldots,S_n)} \;\;
\approx\\
&\approx& 	P(T) \cdot \prod_{i=1}^{n} \frac{P(S_i \mid T)}{P(S_i)} \;\;=\\
&=&		P(T) \cdot \prod_{i=1}^{n} \frac{P(T) \cdot P(S_i \mid T)}
						{P(T) \cdot P(S_i)} \;\;=\\
&=&	P(T) \cdot \prod_{i=1}^{n} \frac{P(T \mid S_i)}{P(T)}
\end{eqnarray*}

In standard statistical part-of-speech tagging there are only two
information sources --- the lexical probabilities and the tags assigned
to neighbouring words. We thus have:
\begin{eqnarray*}
\lefteqn{P(\mbox{Tag $\mid$ Lexicon and $n$-grams}) \;\; =}\\
&=& \frac{P(\mbox{Tag $\mid$ Lexicon}) \cdot P(\mbox{Tag $\mid$
N-grams})}{P(\mbox{Tag})}
\end{eqnarray*}

The context will in general not be fully disambiguated. Rather than
employing dynamic programming over the lattice of remaining candidate
tags, the new approach uses the weighted average over the remaining
candidate tags  to estimate the probabilities:
\begin{eqnarray*}
\lefteqn{P(T \mid \cup_{i=1}^n C_i) \;\;=}\\
&=& \sum_{i=1}^{n}
P(T \mid C_i) \cdot P(C_i \mid \cup_{i=1}^n C_i)
\end{eqnarray*}
It is assumed that $\{C_i : i = 1,\ldots,n\}$ constitutes a partition of
the context $C$, i.e., that $C = \cup_{i=1}^n C_i$ and that $C_i \cap
C_j = \emptyset$ for $i \ne j$. In particular, trigram probabilities are
combined as follows:
\begin{eqnarray*}
\lefteqn{P(T \mid C) \;\;=}\\
&=& \sum_{(T_l,T_r) \in C}
P(T \mid T_l,T_r) \cdot P((T_l,T_r) \mid C)
\end{eqnarray*}
Here $T$ denotes a candidate tag of the current word, $T_l$ denotes a
candidate tag of the immediate left neighbour, and $T_r$ denotes a
candidate tag of the  immediate right neighbour. $C$ is the set of
ordered pairs $(T_l,T_r)$ drawn from the set  of candidate tags of the
immediate neighbours. $P(T \mid T_l,T_r)$ is the symmetric trigram
probability.

The tagger is reductionistic since it repeatedly removes low-probability
candidate tags.  The probabilities are then recalculated, and the
process terminates when  the probabilities have stabilized and no more
tags can be removed without  jeopardizing the recall; candidate tags are
only removed if their probabilities are below some threshold value.

% -------------------------------------------------------------- Sparse Data

\section{Sparse Data}
\label{Sparse}

Handling sparse data consists of two different tasks:
\begin{enumerate}
\item	Estimating the probabilities of events that do not occur in the
	training data.
\item	Improving the estimates of conditional probabilities where the
	number of observations under this conditioning is small.
\end{enumerate}
Coping with unknown words, i.e., words not encountered in the training
set, is an archetypical example of the former task. Estimating
probability distributions conditional on small contexts is an example of
the latter task. We will examine several approaches to these tasks.

For the HMM, it is necessary to avoid zero probabilities. The most
straight-forward strategy is employing the expected-likelihood estimate
(ELE), which simply adds 0.5 to each frequency count and then constructs
a maximum-likelihood estimate (MLE), (see e.g. \cite{Gale90}). The MLE
of the probability is the relative frequency $r$. Another possibility is
the Good-Turing method \cite{Good53}, where each frequency $f$ is
replaced by $f^{\ast} = (f+1) N_{f+1}/N_f$, where $N_f$ denotes the
frequency of frequency $f$.  Alternatively, one can use linear
interpolation of the probabilities obtained by MLE,
$\hat{P}(c \mid a,b) = \lambda_1 r(c) + \lambda_2 r(c \mid b) +
\lambda_3 r(c \mid a,b)$. \cite{Brown92} let the $\lambda$ values
dependent on the context, which improves the tagging accuracy. This is
related to the idea of successive abstraction presented in
Section~\ref{SuccAbs}. To achieve improved estimates of lexical
probabilities, words can be clustered together, see \cite{CuttingEA:92}.

There are several ways to handle unknown words. These include:
\begin{enumerate}
\item   Making every tag a possible tag for that word with equal probability
        and finding the most probable tag solely based on context
        probabilities. The results can be slightly improved by trying only
	open-class tags for unknown words.
\item   As an extension to case 1, choosing different but again constant
        probabilities for each possible tag. This constitutes an a priori
        distribution for unknown words, reflecting for example that most of
        the unknown words are nouns. The probabilities could be obtained
	from a separate training part, or from the distribution of words
	that occur only once in the training corpus. These words reflect
	the distribution of unknown
	words according to the formula presented in \cite{Good53}.
\item   Exploiting word-form information as proposed in \cite{Samuelsson:94}.
        Here, the probability distributions are determined from the last
	$n$ characters of the word, and the remaining number of syllables.
	This method has been proven successful for Swedish text.
\item	Utilizing orthographical cues such as capitalization.
\end{enumerate}

\subsection{Successive Abstraction}
\label{SuccAbs}

Assume that we want to estimate the probability $P(E \mid C)$ of the
event  $E$ given a context $C$ from the number of times $N_E$ it occurs
in $N = |C|$  trials, but that this data is sparse. Assume further that
there is abundant  data in a more general context $C' \supset C$ that we
want to use to  get a better estimate of $P(E \mid C)$.

If there is an obvious linear order  $C = C_m \subset C_{m-1}  \subset
\cdots  \subset C_1 = C'$ of the various  generalizations $C_k$ of $C$,
we can build the estimates of $P(E \mid C_k)$ on the relative frequency
$r(E \mid C_k)$ of event $E$ in context $C_k$ and the previous estimates
of $P(E \mid C_{k-1})$. We call this method {\it linear successive
abstraction\/}. A simple example is estimating the probability  $P(T
\mid l_n,\ldots,l_{n-j})$ of a tag $T$ given $l_{n-j},\ldots,l_n$, the
last $j+1$ letters of the word. In this case, the estimate will be based
on the relative frequencies  $r(T \mid l_n,\ldots,l_{n-j}), r(T \mid
l_n,\ldots,l_{n-j+1}), \ldots, r(T \mid l_n), r(T)$.

Previous experiments \cite{Samuelsson:94} indicate that the following is
a suitable formula:
\begin{eqnarray}
\label{EqLinSuccAbs}
\hat{P}(E \mid C) = \frac{\sqrt{N} \: r(E \mid C) + \hat{P}(E \mid
C')}{\sqrt{N} + 1}
\end{eqnarray}
This formula simply up-weights the relative frequency $r$ by a factor
$\sqrt{N}$,  the square root of the size of context $C$, which is the
active ingredient of the standard deviation of $r$.

If there is only a partial order of the various generalizations, the
scheme is  still viable. For example, consider generalizing symmetric
trigram statistics, i.e., statistics  of the form $P(T \mid T_l,T_r)$.
Here, both $T_l$ and $T_r$ are one-step generalizations of the context
$T_l,T_r$, and both have in turn the common  generalization $\Omega$. We
modify Equation~\ref{EqLinSuccAbs} accordingly:
{\small
\begin{eqnarray*}
\lefteqn{\hat{P}(T \mid T_l,T_r) \;\;=}\\
&=& \frac{\sqrt{|T_l,T_r|} \: r(T | T_l,T_r) + \hat{P}(T | T_l) +
	\hat{P}(T | T_r)}{\sqrt{|T_l,T_r|} + 2}
\end{eqnarray*}
} % \small
and
{\small
\begin{eqnarray*}
\hat{P}(T \mid T_l) &=& \frac{\sqrt{|T_l|} \: r(T \mid T_l) +
\hat{P}(T)}{\sqrt{|T_l|} + 1}\\
\hat{P}(T \mid T_r) &=& \frac{\sqrt{|T_r|} \: r(T \mid T_r) +
\hat{P}(T)}{\sqrt{|T_r|} + 1}
\end{eqnarray*}
} % \small
We call this {\it partial successive abstraction\/}.

% ---------------------------------------------------------------- Experiments

\section{Experiments}
\label{Exp}

For the experiments, both corpora were divided into three sets, one
large set and two small sets. We used three different divisions into
training and testing sets. First, all three sets were used for both
training and testing. In the second and third case, training and test
sets were disjoint, the large set and one of the small sets were used
for training, the remaining small set was used for testing. As a
baseline to indicate what is gained by taking the context into account,
we performed an additional set of experiments that used lexical
probabilities only, and ignored the context.

% --------------------------------------------- Experiments ----- HMM Approach

\subsection{HMM Approach}

The experiments of this section were performed with a trigram tagger as
described in Section~\ref{HMM}. Zero frequencies were avoided by
using expected-likelihood estimation. Unknown words were handled by a
mixture of methods 2 and 3 listed in Section~\ref{Sparse}: If the suffix
of 4 characters (3 characters for the Susanne corpus) of the unknown
words was found in the lexicon, the tag distribution for that suffix was
used. Otherwise we used the distribution of tags for words that occurred
only once in the training corpus.

As opposed to trigram tagging, lexical tagging ignores context
probabilities and is based solely on lexical probabilities. Each word is
assigned its most frequent tag from the training corpus. Unknown words
were assigned the most frequent tag of words that occurred exactly once
in the training corpus. The most frequent tags for single occurrence
words are for the Teleman corpus {\sf NNSS} (indefinite noun-noun
compound) and {\sf noun} (large and small tagset, resp.), for the
Susanne corpus {\sf NN2} (plural common noun) and {\sf NN} (common noun;
again large and small tagset resp.).

Tagging speed was generally between 1000 and 2000 words per second on a
SparcServer 1000; most of this variation was due to variations in the
number of unknown words.

The results for the Teleman corpus are shown in
Table~\ref{TelemanExperimentTable} and the results for the Susanne
corpus in Table~\ref{SusanneExperimentTable}.

\begin{table}
\hrule
\caption{Results of the HMM experiments with the Teleman corpus}
\label{TelemanExperimentTable}
\begin{center}
\tabcolsep=5pt
\begin{tabular}{llll|ccc}
 & & Training	& Testing
	& total correct	& known correct	& unknown correct \\
\hline
\hline
 & & \multicolumn{2}{l|}{Lexical Tagging}\\
 & & A, B, C	& A, B, C	& 95.13\% 	& 95.13\%	& ---     \\
 & & A, B	& C		& 89.27\%	& 94.18\%	& 58.35\% \\
 & & A, C	& B		& 90.42\%	& 94.20\%	& 69.60\% \\
\cline{3-7}
 & & \multicolumn{2}{l|}{Trigram Tagging}\\
 & & A, B, C	& A, B, C	& 96.22\%	& 96.22\%	& ---	  \\
 & & A, B	& C		& 92.88\%	& 94.51\%	& 82.55\% \\
\raisebox{.5ex}[0pt][0pt]{\small\shortstack{S\\m\\a\\l\\l\\
\\T\\a\\g\\s\\e\\t}}
 &
\raisebox{4ex}[0pt][0pt]{\small\shortstack{1\\9\\ \\T\\a\\g\\s}}
 & A, C		& B		& 92.81\%	& 94.62\%	& 82.83\% \\
\hline
\hline
 & & \multicolumn{2}{l|}{Lexical Tagging}\\
 & & A, B, C	& A, B, C	& 90.65\% 	& 90.65\%	& ---     \\
 & & A, B	& C		& 78.84\%	& 89.07\%	& 14.44\% \\
 & & A, C	& B		& 78.05\%	& 88.20\%	& 22.03\% \\
\cline{3-7}
 & & \multicolumn{2}{l|}{Trigram Tagging}\\
 & & A, B, C	& A, B, C	& 98.35\%	& 98.35\%	& ---	  \\
 & & A, B	& C		& 83.78\%	& 89.99\%	& 44.66\% \\
\raisebox{1ex}[0pt][0pt]{\small\shortstack{L\\a\\r\\g\\e\\ \\T\\a\\g\\s\\e\\t}}
 &
\raisebox{4ex}[0pt][0pt]{\small\shortstack{2\\5\\8\\ \\T\\a\\g\\s}}
 & A, C		& B		& 81.01\%	& 89.40\%	& 34.69\% \\
\end{tabular}
\end{center}
\hrule
\end{table}

\begin{table}
\hrule
\caption{Results of the HMM experiments with the Susanne corpus}
\label{SusanneExperimentTable}
\begin{center}
\tabcolsep=5pt
\begin{tabular}{llll|ccc}
 & & Training	& Testing
	& total correct	& known correct	& unknown correct \\
\hline
\hline
 & & \multicolumn{2}{l|}{Lexical Tagging}\\
 & & A, B, C	& A, B, C	& 95.28\% 	& 95.28\%	& ---     \\
 & & A, B	& C		& 91.48\%	& 94.80\%	& 49.72\% \\
 & & A, C	& B		& 91.20\%	& 94.44\%	& 38.37\% \\
\cline{3-7}
 & & \multicolumn{2}{l|}{Trigram Tagging}\\
 & & A, B, C	& A, B, C	& 98.65\%	& 98.65\%	& ---	  \\
 & & A, B	& C		& 95.76\%	& 96.95\%	& 80.81\% \\
\raisebox{.5ex}[0pt][0pt]{\small\shortstack{S\\m\\a\\l\\l\\
\\T\\a\\g\\s\\e\\t}}
 &
\raisebox{4ex}[0pt][0pt]{\small\shortstack{6\\2\\ \\T\\a\\g\\s}}
 & A, C		& B		& 95.18\%	& 96.58\%	& 72.29\% \\
\hline
\hline
 & & \multicolumn{2}{l|}{Lexical Tagging}\\
 & & A, B, C	& A, B, C	& 93.98\% 	& 93.98\%	& ---     \\
 & & A, B	& C		& 86.98\%	& 93.04\%	& 10.78\% \\
 & & A, C	& B		& 88.16\%	& 92.59\%	& 15.81\% \\
\cline{3-7}
 & & \multicolumn{2}{l|}{Trigram Tagging}\\
 & & A, B, C	& A, B, C	& 99.80\%	& 99.80\%	& ---	  \\
 & & A, B	& C		& 92.61\%	& 95.66\%	& 54.20\% \\
\raisebox{1ex}[0pt][0pt]{\small\shortstack{L\\a\\r\\g\\e\\ \\T\\a\\g\\s\\e\\t}}
 &
\raisebox{4ex}[0pt][0pt]{\small\shortstack{4\\2\\4\\ \\T\\a\\g\\s}}
 & A, C		& B		& 93.07\%	& 95.46\%	& 53.83\% \\
\end{tabular}
\end{center}
\hrule
\end{table}

What immediately attracts attention is the remarkably low performance of
the trigram approach for the Teleman corpus. Already the baseline
obtained by lexical tagging is below 80\% for new text, usual results
are around 90\%. Normal results can be obtained only for known words or
when using the small tagset, the latter being in fact a very simple
task,  since the algorithm has to choose from only 19 tags.  For the
large tagset, trigram tagging achieves only 83\% accuracy.  This low
figure is due to the unusually high number of unknown words and  the
larger degree of ambiguity compared to English corpora, as is discussed
in Section~\ref{Teleman}. Using a large Swedish lexicon or morphological
 analyzer should improve the results significantly.

Another interesting result is that accuracy increases when the size of
the tagset increases for the cases where known text is tagged and
context probabilities are taken into account. This means that the
additional information about the context in the larger tagset is very
helpful for disambiguation, but only when disambiguating known text.
This could arise from the fact that a large number ($> 50\%$) of the
trigrams that occur in the training text occur exactly once. And most of
the possible trigrams do not occur at all (generally more than 90\%,
depending on the size of the tagset). Now, the trigram approach has a
distinct bias to those trigrams that occurred once over those that never
occurred. These happen to be the right ones for known text but not
necessarily for new text, thus the positive effect of a larger tagset
vanishes for fresh text.

The results for the Susanne corpus are similar to those reported in
other publications for (other) English corpora.

% ---------------------------------- Experiments ----- Reductionistic Approach

\subsection{Reductionistic Approach}

The reductionistic statistical tagger described in Section~\ref{RedStat}
was tested on the same data as the HMM tagger.  The information sources
employed in the experiments were lexical statistics and contextual
information, which consisted of symmetric trigram statistics. Unknown
words were handled by creating a decision tree of the four last letters
from words with three or less occurrences.  Each node in the tree was
associated with a probability distribution (over the  tagset) extracted
from these words, and the probabilities were smoothened through linear
successive abstraction, see Section~\ref{SuccAbs}.

There were two cut-off values for contexts: Firstly, any context with
less than 10 observations was discarded. Secondly, any context where the
probability distributions did not differ substantially from the
unconditional one was also discarded. Only the remaining ones were used
for disambiguation.  Due to the computational model employed, omitted
contexts are equivalent to backing off to whatever the current
probability distribution is. The distributions conditional on contexts
are however susceptible to the problem of sparse data. This was handled
using partial successive abstraction as described in
Section~\ref{SuccAbs}.

The results are shown in Tables~\ref{TelemanRedExperimentTable}
and~\ref{SusanneRedExperimentTable}. They clearly indicate that:
\begin{itemize}
\item	The employed treatment of unknown words is quite effective.
\item	Using contextual information, i.e., trigrams, improves tagging
	accuracy.
\item	The performance is on pair with the HMM tagger and comparable to
	state-of-the-art statistical part-of-speech taggers.
\item	Teleman is a considerably tougher nut to crack than Susanne.
\end{itemize}
The results using the Susanne corpus are similar to those reported for
the  Lancaster-Oslo-Bergen (LOB) corpus in \cite{deMarcken:90}, where a
statistical $n$-best-path approach was employed to trade precision for
recall.

The tagging speed was typically a couple of hundred words per second on
a SparcServer 1000, but varied with the size of the tagset and the
amount of remaining ambiguity.

\begin{table}
\hrule
\caption{Results of the reductionistic experiments with the Teleman corpus}
\label{TelemanRedExperimentTable}
\begin{center}
\tabcolsep=4pt
\begin{tabular}{@{\hspace{0pt}}c@{\hspace{2pt}}c|l@{\hspace{4pt}}rrrrrrrr}
Training&Testing&Threshold:
	&0.00	&0.05	&0.075	&0.10	&0.15	&0.20	&0.30	&0.50\\
\hline
\hline
\multicolumn{11}{c}{Small Tagset}\\
\hline
& & &\multicolumn{8}{c}{Trigram and lexical statistics}\\
& &Recall (\%)	&100.00	&99.02	&98.66	&98.35	&97.78	&97.37	&96.65	&95.55\\
& &Tags/word	&2.38	&1.15	&1.12	&1.10	&1.07	&1.05	&1.03	&1.00\\
\cline{3-11}
A,B,C	&A,B,C	&	&\multicolumn{8}{c}{Lexical statistics only}\\
& &Recall (\%)	&100.00	&98.96	&98.53	&98.29	&97.69	&97.28	&96.36	&95.10\\
& &Tags/word	&2.38	&1.25	&1.17	&1.14	&1.09	&1.07	&1.03	&1.00\\
\hline
& & &\multicolumn{8}{c}{Trigram and lexical statistics}\\
& &Recall (\%)	&98.98	&97.72	&97.25	&96.81	&96.20	&95.53	&94.67	&93.34\\
& &Tags/word	&2.54	&1.21	&1.17	&1.14	&1.10	&1.07	&1.04	&1.00\\
\cline{3-11}
A,B	&C	&		&\multicolumn{8}{c}{Lexical statistics only}\\
& &Recall (\%)	&98.98	&97.61	&97.14	&96.87	&96.15	&95.63	&94.26	&92.55\\
& &Tags/word	&2.54	&1.34	&1.25	&1.21	&1.14	&1.11	&1.04	&1.00\\
\hline
& & &\multicolumn{8}{c}{Trigram and lexical statistics}\\
& &Recall (\%)	&98.99	&97.80	&97.44	&96.94	&96.34	&95.84	&98.81	&93.50\\
& &Tags/word	&2.51	&1.23	&1.18	&1.15	&1.11	&1.08	&1.04	&1.00\\
\cline{3-11}
A,C	&B	&		&\multicolumn{8}{c}{Lexical statistics only}\\
& &Recall (\%)	&98.99	&97.67	&97.33	&97.07	&96.45	&95.84	&94.34	&92.52\\
& &Tags/word	&2.51	&1.34	&1.26	&1.21	&1.14	&1.10	&1.04	&1.00\\
\hline
\hline
\multicolumn{11}{c}{Large Tagset}\\
\hline
& & &\multicolumn{8}{c}{Trigram and lexical statistics}\\
& &Recall (\%)	&100.00	&98.36	&97.92	&97.54	&97.03	&96.41	&95.31	&93.75\\
& &Tags/word	&3.69	&1.23	&1.18	&1.15	&1.11	&1.08	&1.04	&1.00\\
\cline{3-11}
A,B,C	&A,B,C	&		&\multicolumn{8}{c}{Lexical statistics only}\\
& &Recall (\%)	&100.00	&98.30	&97.63	&97.20	&96.67	&95.57	&93.65	&90.59\\
& &Tags/word	&3.69	&1.43	&1.31	&1.26	&1.22	&1.16	&1.08	&1.00\\
\hline
& & &\multicolumn{8}{c}{Trigram and lexical statistics}\\
& &Recall (\%)	&97.46	&94.93	&93.94	&93.35	&92.35	&91.15	&88.53	&85.56\\
& &Tags/word	&4.16	&1.47	&1.37	&1.31	&1.24	&1.18	&1.08	&1.00\\
\cline{3-11}
A,B	&C	&		&\multicolumn{8}{c}{Lexical statistics only}\\
& &Recall (\%)	&97.46	&95.23	&94.24	&93.69	&92.93	&91.51	&87.92	&83.62\\
& &Tags/word	&4.16	&1.69	&1.53	&1.44	&1.34	&1.26	&1.11	&1.00\\
\hline
& & &\multicolumn{8}{c}{Trigram and lexical statistics}\\
& &Recall (\%)	&96.64	&94.04	&93.00	&92.09	&90.92	&89.46	&86.94	&83.58\\
& &Tags/word	&4.18	&1.48	&1.38	&1.32	&1.24	&1.18	&1.08	&1.00\\
\cline{3-11}
A,C	&B	&		&\multicolumn{8}{c}{Lexical statistics only}\\
& &Recall (\%)	&96.64	&94.51	&93.27	&92.50	&91.02	&89.68	&85.86	&81.69\\
& &Tags/word	&4.18	&1.71	&1.54	&1.44	&1.34	&1.24	&1.10	&1.00\\
\end{tabular}
\end{center}
\hrule
\end{table}

\begin{table}
\hrule
\caption{Results of the reductionistic experiments with the Susanne corpus}
\label{SusanneRedExperimentTable}
\begin{center}
\tabcolsep=4pt
\begin{tabular}{@{\hspace{0pt}}c@{\hspace{2pt}}c|l@{\hspace{4pt}}rrrrrrrr}
%\begin{tabular}{cc|lrrrrrrrr}
Training&Testing&Threshold:
	&0.00	&0.05	&0.075	&0.10	&0.15	&0.20	&0.30	&0.50\\
\hline
\hline
\multicolumn{11}{c}{Small Tagset}\\
\hline
& &		&\multicolumn{8}{c}{Trigram and lexical statistics}\\
& &Recall (\%)	&100.00	&99.46	&99.35	&99.23	&99.03	&98.82	&98.43	&97.75\\
& &Tags/word	&2.07	&1.08	&1.07	&1.06	&1.04	&1.03	&1.02	&1.00\\
\cline{3-11}
A,B,C	&A,B,C	&		&\multicolumn{8}{c}{Lexical statistics only}\\
& &Recall (\%)	&100.00	&99.33	&99.20	&98.94	&98.67	&98.10	&97.43	&95.28\\
& &Tags/word	&2.07	&1.18	&1.16	&1.14	&1.11	&1.08	&1.05	&1.00\\
\hline
& &		&\multicolumn{8}{c}{Trigram and lexical statistics}\\
& &Recall (\%)	&99.22	&98.43	&98.28	&98.11	&97.78	&97.43	&96.91	&95.99\\
& &Tags/word	&2.23	&1.14	&1.11	&1.09	&1.07	&1.05	&1.02	&1.00\\
\cline{3-11}
A,B	&C	&		&\multicolumn{8}{c}{Lexical statistics only}\\
& &Recall (\%)	&99.22	&98.27	&98.03	&97.78	&97.45	&96.80	&96.15	&93.42\\
& &Tags/word	&2.23	&1.25	&1.23	&1.19	&1.15	&1.11	&1.08	&1.00\\
\hline
& &		&\multicolumn{8}{c}{Trigram and lexical statistics}\\
& &Recall (\%)	&99.22	&98.46	&98.22	&97.99	&97.58	&97.15	&96.49	&95.54\\
& &Tags/word	&2.17	&1.13	&1.10	&1.09	&1.06	&1.05	&1.02	&1.00\\
\cline{3-11}
A,C	&B	&		&\multicolumn{8}{c}{Lexical statistics only}\\
& &Recall (\%)	&99.22	&98.21	&97.88	&97.61	&97.35	&96.47	&95.46	&92.87\\
& &Tags/word	&2.17	&1.24	&1.21	&1.17	&1.15	&1.10	&1.06	&1.00\\
\hline
\hline
\multicolumn{11}{c}{Large Tagset}\\
\hline
& &		&\multicolumn{8}{c}{Trigram and lexical statistics}\\
& &Recall (\%)	&100.00	&99.25	&99.12	&98.96	&98.74	&98.44	&98.04	&96.87\\
& &Tags/word	&2.61	&1.10	&1.08	&1.07	&1.06	&1.04	&1.03	&1.00\\
\cline{3-11}
A,B,C	&A,B,C	&		&\multicolumn{8}{c}{Lexical statistics only}\\
& &Recall (\%)	&100.00	&99.05	&98.88	&98.59	&98.20	&97.58	&96.72	&93.98\\
& &Tags/word	&2.61	&1.23	&1.20	&1.17	&1.14	&1.10	&1.07	&1.00\\
\hline
& &		&\multicolumn{8}{c}{Trigram and lexical statistics}\\
& &Recall (\%)	&98.31	&96.94	&96.52	&96.19	&95.68	&95.02	&94.21	&92.70\\
& &Tags/word	&3.01	&1.22	&1.18	&1.15	&1.11	&1.08	&1.04	&1.00\\
\cline{3-11}
A,B	&C	&		&\multicolumn{8}{c}{Lexical statistics only}\\
& &Recall (\%)	&98.31	&96.91	&96.49	&95.94	&95.50	&94.40	&93.42	&90.26\\
& &Tags/word	&3.01	&1.41	&1.35	&1.28	&1.20	&1.14	&1.08	&1.00\\
\hline
& &		&\multicolumn{8}{c}{Trigram and lexical statistics}\\
& &Recall (\%)	&98.49	&97.03	&96.72	&96.41	&95.88	&95.16	&94.29	&92.71\\
& &Tags/word	&2.83	&1.21	&1.18	&1.15	&1.11	&1.08	&1.04	&1.00\\
\cline{3-11}
A,C	&B	&		&\multicolumn{8}{c}{Lexical statistics only}\\
& &Recall (\%)	&98.49	&96.95	&96.55	&96.05	&95.57	&94.44	&93.26	&90.31\\
& &Tags/word	&2.83	&1.36	&1.31	&1.25	&1.19	&1.13	&1.08	&1.00\\
\end{tabular}
\end{center}
\hrule
\end{table}

% ---------------------------------------------------------------- Conclusions

\section{Conclusions}
\label{Conclusions}

The experiments with the HMM approach show that it is much harder to
process the Swedish than the English corpus. Although the two corpora
are not fully comparable because of the differences in size and tagsets
used, they reveal a strong tendency. The difficulty in processing is
mostly due to the rather large number of unknown words in the Swedish
corpus and the higher degree of ambiguity despite having smaller
tagsets. These effects mainly arise from the higher morphological
variation of Swedish which calls for additional strategies to be
applied. These could be the use of a large corpus-independent lexicon
and a separate morphological analysis.

It is reassuring to see that the reductionistic tagger performs as well
as the HMM tagger, indicating that the new framework is as powerful as
the conventional one when using strictly conventional information
sources. The new framework also enables using the same sort of
information as the highly successful Constraint Grammar approach, and
the hope is that the addition of further information sources can advance
state-of-the-art performance of statistical taggers.

Viewed as an extension of the Constraint Grammar approach, the new
scheme allows making decisions on the basis of not fully disambiguated
portions of the input string. The absolute value of the probability of
each tag can be used as a quantitative measure of when to remove a
particular candidate tag and when to leave in the ambiguity.  This
provides a tool to control the tradeoff between recall (accuracy) and
precision (remaining ambiguity).

% ----------------------------------------------------------- Acknowledgements

\section*{Acknowledgements}

We wish to thank Bj\"orn Gamb\"ack for providing information on previous
work with the Teleman corpus.

% --------------------------------------------------------------- Bibliography


\begin{thebibliography}{}

\bibitem[Baum 1972]{Baum72}\ \\
L.~E. Baum.
``An inequality and associated maximization technique in statistical
  estimation for probabilistic functions of Markov processes'',
{\it Inequalities III\/}, pp.\ 1--8, 1972.

\bibitem[Brown {\it et al\/} 1992]{Brown92}\ \\
P.~F. Brown, V.~J.~Della Pietra, F.~Jelinek, J.~D. Lafferty, R.~L. Mercer and
  P.~S. Roossin.
``Class-based $n$-gram models of natural language'',
{\it Computational Linguistics 18(4)\/} pp.\ 467--479, 1992.

\bibitem[Cutting 1994]{Cutting:94}\ \\
Douglass Cutting.
``A Practical Part-of-Speech Tagger'',
in {\it Procs.\ 9th Scandinavian Conference on Computational Linguistics\/},
pp.\ 65--70, Stockholm University, 1994.

\bibitem[Cutting {\it et al\/} 1992]{CuttingEA:92}
Douglass R.\ Cutting, Julian Kupiec, Jan Pedersen and Penelope Sibun.
``A Practical Part-of-Speech Tagger''.
in {\it Procs.\ 3rd Conference on Applied Natural Language Processing\/},
pp.\ 133--140, ACL, 1992.

\bibitem[Eineborg \& Gamb\"ack 1994]{Eineborg;Gamback:94}\ \\
Martin Eineborg and Bj\"orn Gamb\"ack.
``Tagging Experiments Using Neural Networks'',
in {\it Procs.\ 9th Scandinavian Conference on Computational Linguistics\/},
pp.\ 71--82, Stockholm University, 1994.

\bibitem[Francis \& Kucera 1982]{Francis82}\ \\
N.~W. Francis and H.~Kucera.
{\it Frequency Analysis of English Usage\/},
Houghton Mifflin, Boston, 1982.

\bibitem[Gale \& Church 1990]{Gale90}\ \\
W.~A. Gale and K.~W. Church.
``Poor Estimates of Context are Worse than None'',
in {\it Proc.\ of the Speech and Natural Language Workshop\/},
pp.\ 283--287, Morgan Kaufmann, 1990.

\bibitem[Good 1953]{Good53}\ \\
I.~J. Good.
``The population frequencies of species and the estimation of population
parameters'',
{\it Biometrika 40\/}, pp.\ 237--264, 1953.

\bibitem[Karlsson {\it et al\/}\ 1995]{KarlssonEA:95}\ \\
Fred Karlsson, Atro Voutilainen, Juha Heikkil\"a and Arto Anttila (eds).
{\it Constraint Grammar. A Language-Independent System for Parsing
Unrestricted Text\/},
Mouton de Gruyter, Berlin / New York, 1995.

\bibitem[de Marcken 1990]{deMarcken:90}\ \\
Carl G.\ de Marcken.
``Parsing the LOB Corpus'',
in {\it Procs.\ 28th Annual Meeting of the Association for Computational
Linguistics\/}, pp.\ 243--251, ACL 1990.

\bibitem[Rabiner 1989]{Rabiner89}\ \\
L.~R. Rabiner.
``A tutorial on hidden Markov models and selected applications in speech
recognition'',
in {\em Proceedings of the IEEE 77(2)\/}, pp.\ 257--285, 1989.

\bibitem[Sampson 1995]{Sampson95}\ \\
Geoffrey Sampson.
{\it English for the Computer\/},
Oxford University Press, Oxford, 1995.

\bibitem[Samuelsson 1994]{Samuelsson:94}\ \\
Christer Samuelsson.
``Morphological Tagging Based Entirely on Bayesian Inference'',
in {\it Procs.\ 9th Scandinavian Conference on Computational Linguistics\/},
pp.\ 225--238, Stockholm University, 1994.

\bibitem[Teleman 1974]{Teleman:74}\ \\
Ulf Teleman.
{\it Manual f\"or grammatisk beskrivning av talad och skriven svenska\/},
(in Swedish), Studentlitteratur, Lund, Sweden 1974.

\bibitem[Viterbi 1967]{Viterbi67}\ \\
A.~Viterbi.
``Error bounds for convolutional codes and an asymptotically optimum
decoding algorithm'',
in {\it IEEE Transactions on Information Theory}, pp.\ 260--269, 1967.

\bibitem[Voutilainen \& Heikkil\"a 1994]{Voutilainen+Heikkila:94}\ \\
% engcg performance figures with large data:
Atro Voutilainen and Juha Heikkil\"a.
``An English constraint grammar (ENGCG): a surface-syntactic parser of
English'',
in {\it Procs.\ 14th International Conference on English Language
Research on Computerized Corpora\/}, pp.\ 189--199, Z\"urich, 1994.

\end{thebibliography}
\end{document}